\documentclass{article} 
\usepackage{nips14submit_e,times}
\usepackage{url}
\usepackage{wrapfig}
\usepackage{amsmath}
\usepackage{amsfonts}
\usepackage{amssymb}
\usepackage{amsthm}
\usepackage{mathtools}

\usepackage{natbib}

\usepackage[pdftex]{graphicx}
\graphicspath{{img/}}
\DeclareGraphicsExtensions{.pdf,.jpeg,.png}

\usepackage[backref=page]{hyperref} 

\title{End-to-end Continuous Speech Recognition using Attention-based
  Recurrent NN: First Results}

\author{
Jan Chorowski \\
University of Wroc\l{}aw, Poland\\
\texttt{jan.chorowski@ii.uni.wroc.pl} \\
\And
Dzmitry Bahdanau \\
Jacobs University Bremen, Germany
\And
Kyunghyun Cho\\
Universit\'{e} de Montr\'{e}al
\And
Yoshua Bengio \\
Universit\'{e} de Montr\'{e}al \\
CIFAR Senior Fellow
}

\nipsfinalcopy 

\newcommand{\y}{\boldsymbol{y}}
\newcommand{\x}{\boldsymbol{x}}
\newcommand{\h}{\boldsymbol{h}}

\begin{document}

\maketitle


\begin{abstract}
We replace the Hidden Markov Model (HMM) which is traditionally used in in
continuous speech recognition with a bi-directional recurrent neural network
encoder coupled to a recurrent neural network decoder that directly emits a
stream of phonemes. The alignment between the input and output sequences is
established using an attention mechanism: the decoder emits each symbol based on
a context created with a subset of input symbols selected by the attention
mechanism. We report initial results demonstrating that this new
approach achieves phoneme error rates that are comparable to the
state-of-the-art HMM-based decoders, on the TIMIT dataset.
\end{abstract}

\vspace{-3mm}
\section{Introduction}

Speech recognition is challenging because it has to transform a long sequence of
acoustic features into a shorter sequence of discrete symbols such as words or
phonemes. Not only is the problem difficult because the two sequences are
unequal in length, but also because the precise location of the output symbol in
the input sequence is often not known in advance. Therefore, there is no
straightforward way of building a classifier that predicts a target for each
frame of the input signal. The speech recognition model must instead learn to
both align the output sequence to the input sequence and to recognize the content
of the utterance.

The model investigated in this work is a Recurrent Neural Network (RNN) that can
be trained without an explicit alignment between the input and output sequences.
When decoding, the model keeps track of its position in the input sequence
through an attention mechanism. At each step of decoding, the model first scores
all input frames against its hidden state to soft-select relevant input frames.
Next, it summarizes the selection into a context vector and uses it to update the
hidden state and generate the next output symbol. 

This model achieves a phoneme error rate of 18.57\% on the TIMIT
dataset, which is comparable to the state-of-the art DNN-HMM systems but
slightly worse than the best reported error rates obtained using RNNs.
However, it should be noted that this model is very easy to apply: it
requires a narrow beam search, and we found that its accuracy
deteriorates very slightly when greedy search (i.e. taking the most likely
symbol at each time) is used for recognition. It is also easy to implement and tune
(less than a month of work was enough to achieve these results). 

\subsection{Background}

It is possible to obtain the training target for each frame of the acoustic
input by coupling a neural network to a Hidden Markov Model
(HMM)~\citep{Bengio-trnn92,bourlard_connectionist_1994,bengio_neural_1996}. In
this hybrid system the neural network acts as an acoustic model by predicting
the state of the HMM from each corresponding input frame.  
Since a target
is provided for each input frame, we can train the neural network in a usual
way by minimizing the classification error.
Once a per-frame classifier is trained, the whole system including the acoustic
model and the HMM can be tuned jointly on full sentences to minimize the
decoding
error~\citep{lecun_gradient-based_1998,he_discriminative_2008,kingsbury_lattice-based_2009,vesely_sequence-discriminative_2013}.
This hybrid approach has recently made important progress by adopting, as an
accoustic model, deep neural networks such as fully-connected feedforward neural
networks~\citep{a_mohamed_deep_2009,hinton_deep_2012}, convolutional
networks~\citep{sainath_deep_2014}, and recurrent
networks~\citep[RNN,][]{graves_hybrid_2013}.

The hybrid architecture (NN-HMM) is, however, rather complicated and requires
controlling the relative contribution from each part of the model to
the decoding error. Moreover, training needs to be performed in multiple stages,
beginning with a hybrid architecture having Gaussian Mixture Model (GMM) and HMM
which is used to generate per-frame target states (forced-alignment). This is followed by
iteratively training an acoustic model (neural network) and re-estimating the
transition probabilities of the HMM. Furthermore, it has been reported by, for
instance, \citet{graves_hybrid_2013} that the improvements in the per-frame
classification do not necessarily translate to the decoding accuracy.

In contrast, some earlier works have proposed to minimize the final decoding
error directly by optimizing the sum of costs along the paths in an
alignment/decision graph~\citep{Bengio-trnn92,lecun_gradient-based_1998}.  Along
this line of research, \citet{graves_connectionist_2006} proposed, more
recently, an alternative approach called the Connectionist Temporal Classifier
(CTC), which can be used without an explicit input-output alignment. 

In the CTC-based model, each per-frame prediction is either one of the set of
desired outputs (words, phones) or a special 
separator symbol. The final sequence is obtained by removing the separators and
merging blocks of consecutive identical output symbols. A forward-backward
algorithm is used to exactly compute the probability of a desired output
sequence given the per-frame predictions. 

As an extension of the CTC \citet{graves_sequence_2012} proposed the RNN
Transducer. Unlike the CTC, which can be seen as an acoustic-only model, the RNN
Transducer has another RNN that acts as a language model. The transducer is
composed of two parts: (1) a transcription network that produces for each frame
of the input sequence either a target symbol or the separator and (2) a
prediction network that generates the final sequence of outputs without
separators and repeated symbols. These two components allow the RNN Transducer
to model the probability of observing the next output symbol given its position
in both the input and output sequences. Similarly to the CTC, the probability of
observing an output sequence for a given input is computed using the
forward-backward algorithm.

The RNN Transducer computes the score of each possible output token based on the
position in the input and output sequences. Originally, the score was obtained
by multiplying the separate scores from the transcription and prediction
networks. \citet{graves_speech_2013}, on the other hand, used a separate
multi-layer perceptron (MLP) to combine the two scores from the transcription
and prediction networks and achieved the state-of-the-art performance on the
TIMIT dataset.

\subsection{Attention Mechanism}

The model we consider in this paper is closely related to the RNN Transducer,
however, with an attention mechanism that decides which input frames be used to
generate the next output element. 

The attention mechanism in this respect was first used by
\citet{graves_generating_2013} to build a neural network that generates
convincing handwriting from a given text. At each step, the
network predicts a \mbox{(soft-)window} over the input sequence 
that corresponds to the character being currently written. 

A similar approach of attention was used more recently in a so-called ``neural
machine translation model''~\citep{bahdanau_neural_2014}. In this case, for
generating each target word, the network computes a score matching the hidden
state of an output RNN to each location of the input sequence
\citep{bahdanau_neural_2014}. The scores are normalized to sum to one over the
input sequence and can be interpreted as a probability of each input location
being aligned to the currently generated target word.

Unlike the handwriting generation network, the attention mechanism in the neural
translation model may assign high scores to non-adjacent locations in the input
sequence, which allows it to perform long-distance word reordering. In addition,
it uses the features of the input sequence extracted by a bidirectional
recurrent neural network, and not just the output RNN state, in order to predict
these alignment probabilities.

In this paper, we propose a model, based on this neural translation model, that
is more suited to speech recognition. We allow the learned soft alignment procedure
to take the relative position into account and add a penalty helping the
attention mechanism to choose a single, narrow mode and encourage that mode of attention to
move forward. This helps the model to search nearby, potentially near-future,
input frames given the current belief/state about which output symbols have been
generated. The learned dependency on the relative position of successive points of
attention can, further, be used to speed up decoding, as the final
model has to consider only a tiny section of the input sequence when generating
each output symbol.

\section{Model architecture}

\begin{figure}[t]
  \centering
  \includegraphics[width=\textwidth]{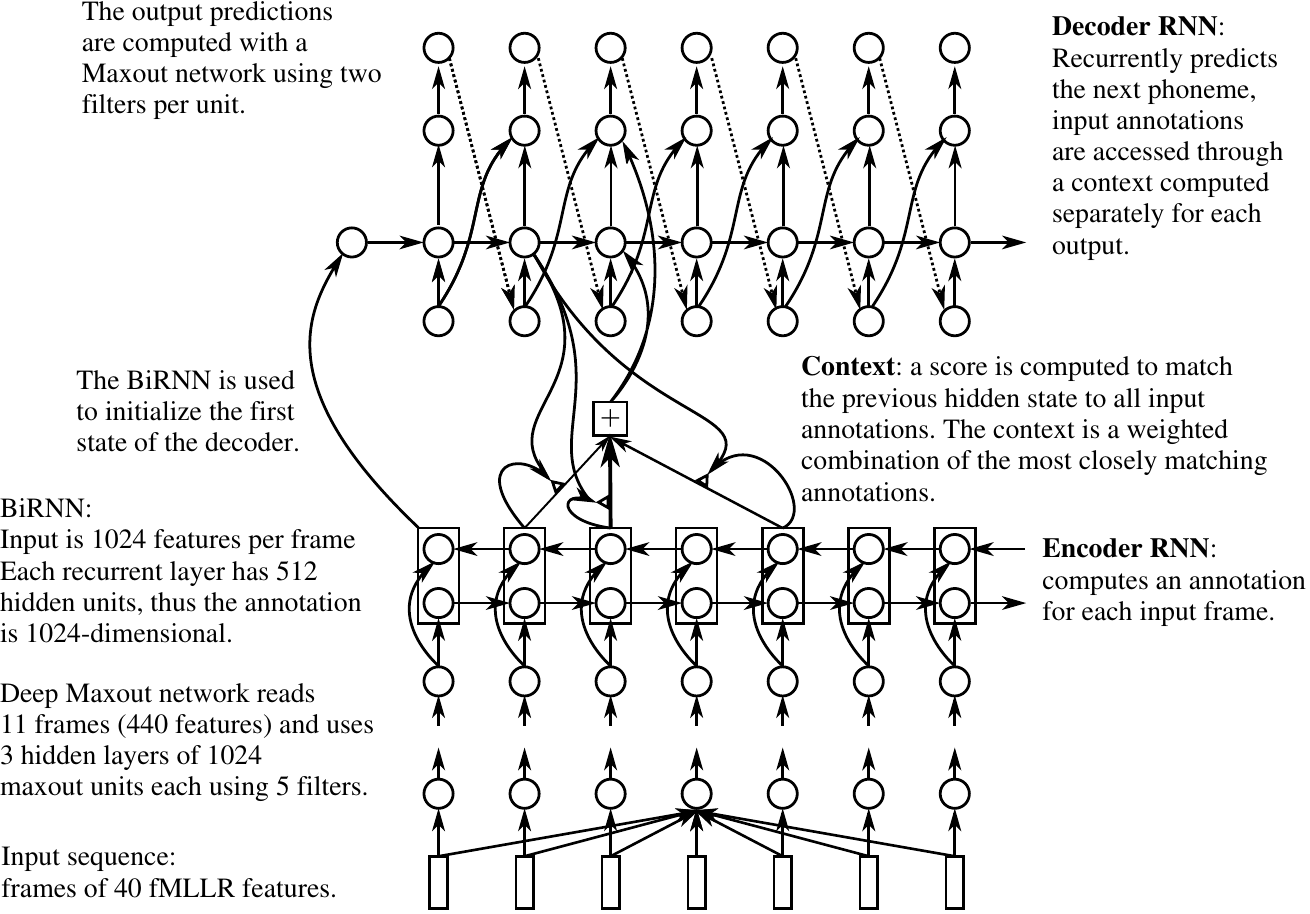}
  \caption[Model Architecture]{Proposed model architecture. The system contains
      three parts: an encoder that computes annotations of input frames
      (learned features that may depend on the whole sequence), an attention
      mechanism that decides where to look in the input sequence to provide a
  context for emitting the next output, and a generative output RNN which
  iteratively predicts the next phoneme conditioned on its state and the
  context. For visual simplicity we have shown only one context computation.} 
    \label{fig:model}
    \vskip -3mm
\end{figure}

The model consists of an \emph{encoder} that maps the raw input
sequence $\x=(x_i;i=1,\ldots,I)$
to a sequence of features $\h=(h_i;i=1,\ldots ,I )$ (also called
annotations), 
a \emph{decoder} that generates the output sequence\footnote{
    The word ``decoder'' here is used in a sense that comes from the
    auto-encoder analogy, with the encoder producing the internal
    representation of the input sequence and the decoder mapping it to the
    output sequence distribution. This is different from the usual use of the word
    decoder in speech recognition systems to talk about the search algorithm
    which approximately looks for the most probable configuration of latent and
    output variables (the output sequence), e.g., using beam-search. We also
    use an approximate search mechanism (with beam search) to look for the most
    probable output sequence, given the input sequence.
} $\y=(y_o; o=1,\ldots ,O)$ and an \emph{attention
mechanism} that matches parts of the input sequence with elements of
the output sequence~\citep{bahdanau_neural_2014}.
A graphical illustration of the model is presented in Fig.~\ref{fig:model}. 

\subsection{Decoder}

\begin{figure}[t]
  \centering
  \begin{tabular}{rl}
    (a) & \includegraphics[width=.9\textwidth]{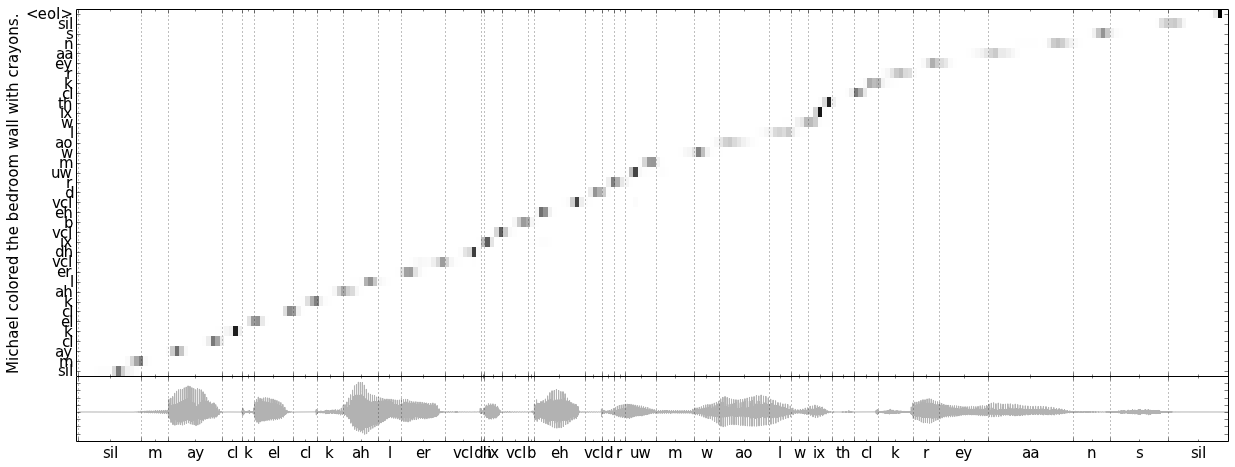}\\
    (b) & \includegraphics[width=.9\textwidth]{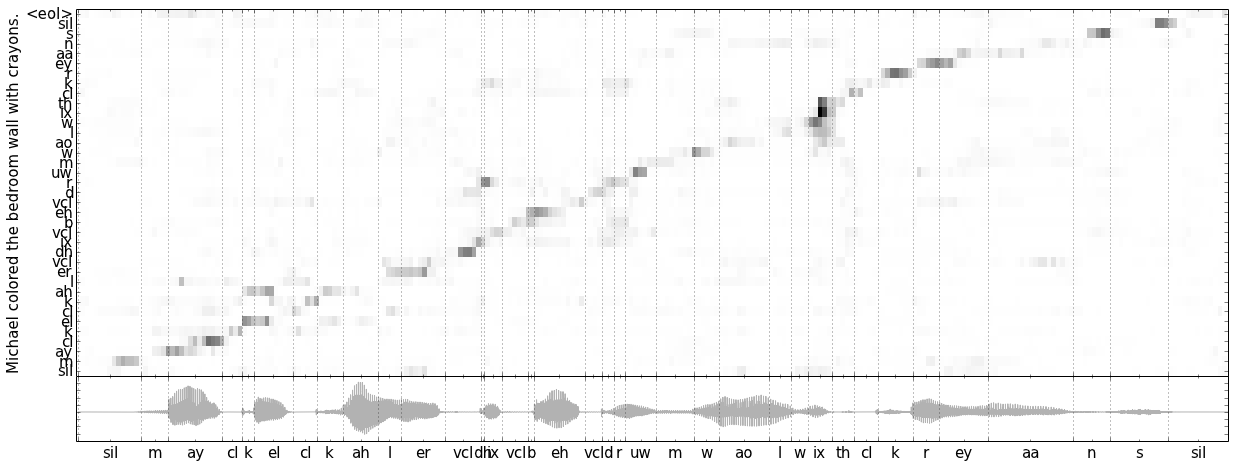}
  \end{tabular}
  \caption[Alignments produced by the model.]{Alignments produced by the model:
      (a) when the alignment was successfully  encouraged to be monotonic, and
      (b) when the model is free to select any frame in the input sequence. Each
      row in the plot contains the scores computed by the attention mechanism
      between the previous hidden state and all the input annotations. 
      In (b), we observe how the absence of the learned
      preference for monotonicity makes the model confused by the
      repeated occurrence of the phonemes ``cl k'', and to a
      lesser degree, by the repetition of ``w''. 
    }
    \label{fig:alignments}
\end{figure}

The decoder predicts the output sequence $\y$ conditioned on the annotations
$\h$ of an input sequence $\x$ by factorizing $p(\y | \h)$ into ordered
conditionals:
\begin{equation*}
  p(\y|\h) = \prod_{o=1}^O p(y_o | y_{o-1}, y_{o-2}, \ldots, y_1, \h)
\end{equation*}
The RNN decoder uses a subset of the input annotations summarized into
a context $c_o$ to compute both the update to its state and the 
prediction of the next output symbol:
\begin{align*}
  s_o &= f(s_{o-1}, y_{o-1}, c_o), \\
  p(y_o|y_{o-1}, y_{o-2}, \ldots, y_1, \h) &= g(y_{o-1}, s_o, c_o), 
\end{align*}
where $s_{o-1}$ and $s_o$ are the previous and current state vectors, while
$y_o$ is the prediction of the next output. The recurrent state
update ($f$) is computed using an affine layer with a reset-update
gate used as the activation function \citep{cho_learning_2014} to keep
track of long-term dependencies. The next output prediction ($g$) is
realized with an MLP composed of a Maxout
\citep{goodfellow_maxout_2013} and SoftMax layers. 

The context is a weighted sum of annotations:
\begin{equation*}
  c_o = \sum_{i=1}^I \alpha_{o,i} h_i,
\end{equation*}
where $\alpha_{o,i}$ is a normalized weight for each annotation $h_i$. This
effectively means that the decoder selects each annotation $h_i$ with a
certainty $\alpha_{o,i}$.

The selection is performed in two steps. First, scores $e_{o,i}$ are computed to
match the previous state of the decoder to all annotations. The scores are then
penalized based on the relative position of the current and previous selection,
and normalized:
\begin{align}
  e_{o,i} &= a(s_{o-1}, h_i) \label{eq:context_scoring}, \\
  \hat{e}_{o,i} &= d(i-\mathbb{E}_{\alpha_{o-1}}[i])\exp(e_{o,i})
  =
  d\left(i-\sum_{k=1}^I\alpha_{o-1,k}k\right)\exp(e_{o,i}), \label{eq:gating-fn} \\
  \alpha_{o,i} &= \frac{\hat{e}_{o,i}}{\sum_{k=1}^I\hat{e}_{o,k}},
  \nonumber
\end{align}
where the computation $a(\cdot, \cdot)$ is an MLP with one hidden layer and
linear output, while  $d(\cdot)$ is an MLP with a single hidden layer and
logistic sigmoid output ($\left[ 0, 1\right]$).

The proposed addition of the gating procedure can be understood
as follows.  The attention mechanism searches through the input
sequence to find frames that match the current state of the
decoder. However, in an utterance there may be repeated phonemes
that have very similar annotations, which consequently result in
matching to all similarly sounding locations, as shown in
Fig.~\ref{fig:alignments}~(b).  The gating procedure prevents
this behavior by confining the search to locations that are near
the inputs relevant to the previously generated symbol. 

\begin{figure}[t]
 \centering
 \begin{minipage}{.49\textwidth}
   \centering
   \includegraphics[width=.95\textwidth]{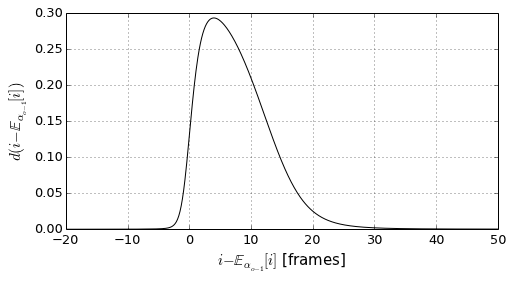}
   \caption{Learned gating function $d(\cdot)$ (Eq.~\ref{eq:gating-fn}).
   }
   \label{fig:gating}
 \end{minipage}\hspace{.01\textwidth}%
 \begin{minipage}{.49\textwidth}
     \includegraphics[width=.92\textwidth]{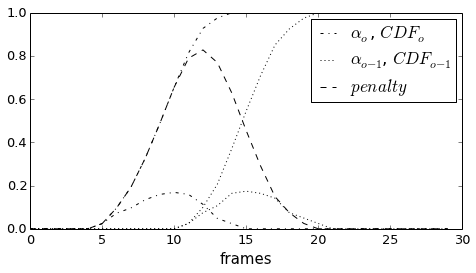}
     \caption[Penalty computation visualization]{
     Illustration of the penalty
     in Eq.~\eqref{eq:ali_monotonicity}.
   }
   \label{fig:alignment_penalty}    
 \end{minipage} 
\end{figure}

It is, however, important to notice that the shape of the gating function $d$ is
not defined a priori, but learned. In Fig.~\ref{fig:gating}, we show the shape
of the learned gating function. $d(\Delta)$ directly encodes the preference of
the input frames $\Delta$ steps away from the previously searched input location
$E_{\alpha_{o-1}}[i]$. We clearly see that the model automatically learned to
prefer approximately 4 frames in the future and strongly inhibit the selection
of any past frames ($\Delta < 0$). Most of the mass is concentrated between
$\Delta=0$ and $12$.

\subsubsection{Learning to prefer monotonic alignments}

We would like to constrain the selection of the frames for consecutive outputs
to advance monotonically in time. To motivate the network to find monotonic
alignments we design a penalty that is added to the optimization cost. 

We penalize any alignment that maps to inputs which were already considered for
output emission. Since the selection weights $\alpha_{o,i}$ are normalized, their
cumulative sum over the input sequence is monotonically increasing and bounded
in the range $[0,1]$.  To encourage the monotonicity of the alignment over time,
we add to the optimization cost the differences $p_o$ between the cumulative
sums of the selection weights used to emit the current and previous phoneme:
\begin{equation} 
    \label{eq:ali_monotonicity}
    p_o = \max\left(0,\sum_{i=1}^I\left(\sum_{k=1}^{i}\alpha_{ok} -
    \sum_{k=1}^i\alpha_{o-1,k} \right)\right)
\end{equation}

An example of the penalty term is shown in
Fig.~\ref{fig:alignment_penalty}. This figure shows the alignment
weights and their CDFs computed for $o-1$ and $o$ (represented by
two dotted and dot-dashed curves), along with the penalty cost
(long dashes). Intuitively, the proposed penalty encourages the
attention mechanism to select nearby locations in the input
sequence that are in the future, rather than in the past with
respect to the previously selected location.

The effect of the proposed penalty term together with the gating
procedure (see Eq.~\eqref{eq:gating-fn} can clearly be observed
in Fig.~\ref{fig:alignments}. In Fig.~\ref{fig:alignments}~(a)
where both the gating procedure and the penalty term were used,
the alignment at each time step is confined to a small subset of
consecutive locations and is monotonic. On the other hand, the
alignment without the proposed penalty shown in
Fig.~\ref{fig:alignments}~(b) clearly does not show these
properties.

Although this regularization term could have a hyper-parameter
coefficient multiplying it, preliminary experiments suggested
that a value of 1 worked well and we did not optimize it further. 

\subsection{Encoder}

The encoder is implemented using a cascade of a multilayer Maxout
network~\citep{goodfellow_maxout_2013} whose last layer is fed to a
bi-directional RNN
(BiRNN)~\citep{schuster_bidirectional_1997,graves_hybrid_2013}. We have chosen
this hybrid architecture to combine the ability of the Maxout network to
nonlinearly transform speech features with the efficient summarization of nearby
preceding and following input frames provided by the BiRNN. The BiRNN
uses the reset-update gate to account for long-term dependencies \citep{cho_learning_2014}.

\subsection{Training objective}

The model is trained to minimize the negative log-likelihood of
observing correct phoneme sequences conditioned on recorded
utterances, with extra penalty terms added to promote alignment
monotonicity \eqref{eq:ali_monotonicity} and with weight decay applied
only to weights of the output MLP $g$.

\section{Experiments}

\subsection{Problem setup}
We evaluated the model on the TIMIT dataset. The
baseline score was established using the recipe ``s5'' of the
Kaldi toolkit \citep{povey_kaldi_2011}, using the ``score\_basic.sh''
scoring script\footnote{Two scoring scripts are provided ({\tt basic} and {\tt sclite}), differing in
the way errors in the silence phone are counted. The {\tt basic} scoring
script treats them as any other error, while the {\tt sclite} scorer can be
configured to disregard errors related to recognizing silence
phones.}. 
All results are gathered in
Table \ref{tb:results}. All models were trained to recognize 48
phonemes, their predictions were converted to the 39 phoneme set
before scoring. All SA sentences were removed from both training and
testing sets. The models were evaluated on the 24 speaker core test
set with an auxiliary 50 speaker development set used to select the
best network. The GMM-HMM and DNN-HMM hybrids used phoneme
bi-grams as the language model used during decoding. The
attention-based recurrent network used no language model.
Similarly to the Kaldi's DNN recipe we kept 10\%
of the training data for validation, which resulted in a final 4-way split of the data
into testing, development, validation, and training sets\footnote{We have thus used the same
data preprocessing as the Deep NN part of the TIMIT s5 recipe.}.

The networks were trained on speaker adapted fMLLR features obtained using the GMM-HMM
built during stage ``tri3'' of the ``s5'' recipe provided by Kaldi. 
Each acoustic frame was described using 40 features, however
the network had access to segments of 11 frames totaling in 440
features per frame. Furthermore we used the ``tri3'' triphone
GMM-HMM model to force-align the data to obtain 
per-frame training targets. 

The networks were implemented using the Theano \citep{bergstra_theano:_2010} and
Pylearn2 \citep{goodfellow_pylearn2:_2013} libraries.

\subsection{RNN training and evaluation}

Due to slow convergence of the training process the experiment was limited
to training a single network, with regularization and training
hyper-parameters changed during the experiment run. The obtained
results should thus be treated as a preliminary proof-of-concept validation of
the proposed model, rather than a rigorous analysis of its
performance.

To ease training of the network we decided to initialize the acoustic part
of the RNN encoder with weights of a deep Maxout network trained with
dropout regularization
\citep{goodfellow_maxout_2013,hinton_improving_2012} to predict the
states of the HMM. The network was trained purely discriminatively
(no layer-wise initialization) to predict per-frame HMM states. The
architecture of the network was mildly optimized, and coupled with an
HMM decoder it reached 18.41\% development set error and 19.68\% test
error.

Next we initialized the proposed RNN encoder-decoder. 
We removed the final affine transformation and softmax layer from the
acoustic network and used it to initialize the acoustic part of the RNN
encoder. Its parameters were then frozen until the  RNN
encoder-decoder reached
good recognition performance. We initialized other weights randomly, however the
weights in the recurrent connections were orthonormalized to ease
training. For the first training iterations the network did not use the gating
mechanism to disambiguate between matching input frames \eqref{eq:gating-fn}.

\begin{table}[t]
\caption{Phoneme error rates of evaluated models}
\label{tb:results}

\vspace{2mm}
\centering
\begin{tabular}{l|l|l}
\multicolumn{1}{c|}{\bf Model}  &\multicolumn{1}{c|}{\bf DEV} &\multicolumn{1}{c}{\bf TEST} \\ 
\hline 
\hline 
\multicolumn{3}{c}{Kaldi TIMIT s5 recipe with {\tt basic} scorer} \\
\hline 
Speaker independent triphone GMM-HMM & 23.15\% & 24.32\% \\
Speaker adapted triphone GMM-HMM & 20.56\% & 21.65\% \\
DBN-HMM with SMBR training & 17.55\% & 18.79\% \\
DBN-HMM with SMBR training and greedy search & 32.02\% & 33.00\% \\
\hline 
\multicolumn{3}{c}{Proposed model with the {\tt basic} scorer} \\
\hline 
Maxout network with per-frame training & 18.41\% & 19.68\%   \\
RNN model with frozen acoustic layer & 17.53\% & 18.68\% \\
RNN model trained end-to-end & 16.88\% & 18.57\% \\
RNN model trained end-to-end with greedy search & 17.06\% & 18.61\%\\
\hline 
\multicolumn{3}{c}{References} \\
\hline 
Deep RNN Transducer \citep{graves_speech_2013}&
\textcolor{white}{0}N/A & 17.7\% \\
\end{tabular}
\end{table}

The RNN was trained with stochastic gradient descent using the
AdaDelta learning rule \citep{zeiler_adadelta:_2012} and an adaptive
gradient clipping mechanism described in section
\ref{sec:training}. To save on computation time, mini-batches were
composed of utterances of similar length (we re-shuffled the data
on each epoch, divided it into groups of up to 32 utterances and sorted
each group before forming the batches). We trained with a
minibatch size 4. Each utterance was processed
with the Maxout network, and then we appended a frame containing only zeros to
indicate to the RNN where the utterance ends.

When the network started to generate good phoneme sequences 
we enabled the gating mechanism and added to the training cost the
penalty for alignment non-monotonicity (c.f. sec. 
\eqref{eq:ali_monotonicity}). The gating MLP ($d$ in
eq. \eqref{eq:gating-fn}) receives large inputs (up to the number of input
frames), therefore its hidden layer was manually initialized: the weights were set to the constant $10^{-3}$,
while the corresponding biases were uniformly distributed in the
$(-5,5)$ range. At the same time we added a weight
decay penalty of $10^{-3}$ applied to the weights of the MLP predicting the next
phoneme and of $2\cdot 10^{-4}$ to the weights in the network used to score input
frames in Eq.~\eqref{eq:context_scoring}. The best
decoding error reached on the development set was 17.53\% which
corresponded to 18.68\% on the test set.

Finally, we included into training the pre-initialized Maxout
part of the encoder. However, we disabled dropout regularization
as we found that it increased the decoding errors on the
development set by a few percentage points and the network was
very slow to recover. The training of the full network lowered
the development set error to 16.88\%, which corresponded to a
test error of 18.57\%.

In an auxiliary experiment we analyzed the importance of the
beam-search width on decoding accuracy. We observed that a narrow beam
width is required to reach optimal performance (decoding with beam
width 10 is sufficient) and that the
accuracy degraded very slightly if the beam serach is replaced
with a greedy search. To put this result into context we have
used the HMM decoder with a beam search width limited to 1. We have
tuned the weighting of the acoustic and language model scores. Both development and test
error rates increased to about 32\%. This result shows is an important
characteristic of the proposed model as it suggests that the RNN
encoder-decoder really learns how to align the input and output sequences.

\subsubsection{Improving the Convergence of AdaDelta}
\label{sec:training}

In order to avoid the issue of exploding gradient, we rescaled
the norm of the gradient, if it went over a predefined
threshold~\citep{pascanu_difficulty_2013}. However, with this
gradient rescaling, we observed that AdaDelta failed to converge
to a lower training error at the later stage of training. We
addressed this problem by multiplying the gradient with a small
scalar (typically, $10^{-2}$) at the later stage of learning.
As this procedure clearly affects the original gradient
rescaling, we used the following algorithm to automatically
adapt the gradient rescaling threshold:
\begin{align*}
    \nabla_\Theta &= \text{gradient of the loss}, g = \nabla_\Theta \cdot \text{gradient scale}, ng = \sqrt{\sum g_i^2} \\
    \text{Elog}_{ng} &= \min\left(\rho,
      \frac{\text{nsteps}}{\text{nsteps+1}}\right) \text{Elog}_{ng}  + 
    \left(1-\min\left(\rho, \frac{\text{nsteps}}{\text{nsteps+1}}\right)\right)\log(ng) \\
    \text{Elog2}_{ng} &= \rho \text{Elog2}_{ng} + (1-\rho) \log^2(ng) \\
    rg &= \begin{cases} g, & \text{ if } ng \leq \exp\left(\text{Elog}_{ng} + \kappa *
      \sqrt{\max\left(0, \text{Elog2}_{ng} - \text{Elog}_{ng}^2\right)}\right) \\ 
      g \frac{\exp(\text{Elog}_{ng})}{ng}, & \text{otherwise}
    \end{cases} 
\end{align*}
The rescaled gradient $rg$ was then passed to the AdaDelta update procedure.

The accumulators $\text{Elog}_{ng}$ and $\text{Elog2}_{ng}$ were
initialized to 0 at the beginning of the run. The algorithm uses
a smaller decay constant to compute the moving average of the
mean in the early stage to underestimate the standard deviation.
This has the effect of having a threshold close to the mean value
at the beginning and increasing the threshold only when the
running averages are correctly tracking the norm of the gradient.

\section{Conclusions}

We report preliminary results of phoneme recognition with an RNN
model that does not require an explicit alignment and produces a
desired output sequence directly, without resorting to
intermediate per-frame predictions that have to be processed with
a specialized decoding algorithm. The proposed model is an extension of
the network used for neural machine
translation~\citep{bahdanau_neural_2014} and is closely related
to CTC and the RNN
Transducer~\citep{graves_connectionist_2006,graves_speech_2013}. 
Although closely related, our model differs from these models in
multiple aspects.

For instance, the proposed approach is related to the RNN
Transducer in the sense that scores are computed
for all pairs of positions in the input and output sentences.
However, in the RNN Transducer the scores define a distribution
over a latent alignment, which is then marginalized out. The proposed model (a
variant of RNN
Encoder--Decoder~\citep{cho_learning_2014,bahdanau_neural_2014})
instead uses the scores as an explicit alignment which is used to
compute a context vector. In addition the decoder state of our model 
contains information about previous alignment choices unlike the 
states of the generation network from the RNN Transducer.

Accessing the input sequence through a context vector is related
to the model that generates handwritten characters, proposed
recently by \citet{graves_generating_2013}.  Their handwriting
generation network, however, predicts the location of the inputs
selected for the next step, while our model scans through all
inputs looking for correctly matching patterns. This scanning of
the whole sequence helps the model deal with irregularities in
speech, such as long pauses. 
Nevertheless, our model learns an
expected distance between inputs pertaining to subsequent output
symbols to disambiguate between occurrences of similar input
frames (see Fig. \ref{fig:gating}). This relation can be used to speed up the recognition by
confining the search for matching frames to the most probable
locations only. However, even without this optimization real-time
decoding is possible. Our beam search decoder achieved a real time
factor of 0.3 with beam width 10 when a GTX480 GPGPU was used to compute
network activations.

We observed that our model performs well even with a simple,
narrow-beam search which is close to greedy search.  This fact
has its significance in extending the proposed approach to a
large vocabulary speech recognition system. As the decoding of
the phoneme stream is nearly deterministic, we may simply put on
top of the phone sequence another RNN that models a sequence of
words. In this case, we will be able to directly search for the
most probable sequence of words, instead of the phoneme level or
frame level (as is done by the HMM-based hybrid systems).

\subsubsection*{Acknowledgments}


The authors would like to thank Felix Hill for his feedback
and acknowledge the support of the following agencies for
research funding and computing support: NSERC, Calcul Qu\'{e}bec, Compute Canada,
the Canada Research Chairs and CIFAR. Jan Chorowski was funded under
the project ``Development of the potential and educational offer of
the University of Wroc\l{}aw - the chance to enhance the competitiveness
of the University,'' co-sponsored by the European Social Fund.


\small
\bibliographystyle{natbib}
\bibliography{strings,strings-shorter,ml,aigaion,refs}




\end{document}